\documentclass{article}

\usepackage{arxiv}
\usepackage[USenglish]{babel}
\usepackage[utf8]{inputenc} 
\usepackage[T1]{fontenc}    
\usepackage{hyperref}       
\usepackage{url}            
\usepackage{booktabs}       
\usepackage{amsfonts}       
\usepackage{nicefrac}       
\usepackage{microtype}      
\usepackage{lipsum}		
\usepackage{graphicx} 
\usepackage{doi}
\usepackage{bm}
\usepackage{amsmath}
\usepackage{amssymb,MnSymbol, multirow}
\usepackage{longtable} 
\usepackage{array}
\usepackage{rotating}
\usepackage{color, colortbl}
\definecolor{Gray}{gray}{0.9}
\usepackage{titlesec}
\usepackage[acronym]{glossaries}

\usepackage[misc]{ifsym}
\usepackage{subcaption}
\usepackage{tikz}
\usepackage{algorithm}
\usepackage{algpseudocode}
\usepackage{tabularx}
\usepackage{verbatim}
\usepackage{ragged2e}

\newacronym{drl}{DRL}{Deep Reinforcement Learning}
\newacronym{dl}{DL}{Deep Learning}
\newacronym{rl}{RL}{Reinforcement Learning}
\newacronym{il}{IL}{Imitation Learning}
\newacronym{greedy}{$Greedy_{90\%}$}{Greedy Expert Agent}
\newacronym{ppo}{PPO}{Proximal Policy Optimization}
\newacronym{re}{RE}{Renewable Energy}
\newacronym{ml}{ML}{Machine Learning}
\newacronym{ca}{CAgent}{CurriculumAgent}
\newacronym{pbt}{PBT}{Population Based Training}
\newacronym{cem}{CEM}{Category Embedding Model}
\newacronym{mdp}{MDP}{Markov Decision Process}
\newacronym{sgd}{SGD}{Stochastic Gradient Decent}
\newacronym{tso}{TSO}{Transmission System Operator}
\newacronym{opf}{OPF}{Optimal Power Flow}
\newacronym{l2rpn}{L2RPN}{Learning to Run a Power Network}
\newacronym{gnn}{GNN}{Graph Neural Network}
\newacronym{mst}{MST}{Median Survival Time}
\newacronym{tts}{TTs}{Target Topologies}
\newacronym{tt}{TT}{Target Topology}
\newacronym{mstcm}{MSTCM}{Median Survival Time across Chronic Medians}
\newacronym{sacd}{SACD}{Soft Actor-Critic Discrete }
\newacronym{senior}{$Senior_{95\%}$}{Senior Agent}
\newacronym{topo}{$TopoAgent_{85-95\%}$}{Topology Agent} %
\newacronym{senior85}{$Senior_{85\%}$}{85\%-Senior Agent}
\newacronym{d_n}{$DoNothing$}{Do-Nothing Agent}
\newacronym{smaac}{SMAAC}{Semi Markov Afterstate Actor Critic}
\newacronym{ddqn}{DDQN}{Dueling Deep Q Network}
\newacronym{bohb}{BOHB}{Bayesian Optimization Hyperband}
\newacronym{pca}{PCA}{Principal Component Analysis}
\newacronym{rf}{RF}{Random Forest}
\newacronym{xg}{XGBoost}{Extreme Gradient Boosting}
\newacronym{lgbm}{LightGBM}{Light Gradient-Boosting Machine} 
\newacronym{gandalf}{GANDALF}{Gated Adaptive Network for Deep Automated Learning of Features}
\newacronym{gflu}{GFLU}{Gated Feature Learning Units}
\newacronym{gru}{GRU}{Gated Recurrent Units}
\newacronym{mlp}{MLP}{Multi Layer Perceptron}
\newacronym{ood}{OOD}{out-of-distribution} 
\newacronym{tpe}{TPE}{Tree-structured Parzen Estimator} 
\newacronym{mcts}{MCTS}{Monte Carlo Tree Search}
\newacronym{fnn}{FNN}{Feed-forward Neural Network}
\newacronym{gat}{GAT}{Graph Attention Network}

\title{Learning Topology Actions for Power Grid Control: A Graph-Based Soft-Label Imitation Learning Approach}

\date{} 				

\date{} 				
\author{
    \textbf{Mohamed~Hassouna}\textsuperscript{1,2,}\thanks{Corresponding author at: Fraunhofer Institute for Energy Economics and Energy System Technology (IEE), Joseph-Beuys-Straße 8, Kassel, 34117, Germany. E-mail address: mohamed.hassouna@iee.fraunhofer.de. \url{https://orcid.org/0000-0001-9927-1625}}\and
    Clara~Holzhüter\textsuperscript{1,2}\and
    Malte~Lehna\textsuperscript{1,2} \and
    Matthijs~de~Jong\textsuperscript{3} \and
    Jan Viebahn\textsuperscript{3} \and
    Bernhard Sick\textsuperscript{2} \and
    Christoph~Scholz\textsuperscript{1,2} \and
\\
\textsuperscript{1} Fraunhofer Institute for Energy Economics and Energy System Technology (IEE)\\Joseph-Beuys-Straße 8, Kassel, 34117, Germany\\
\textsuperscript{2} Intelligent Embedded Systems, University of Kassel \\
\textsuperscript{3} TenneT TSO B.V., Arnhem, The Netherlands
\\
\\
}

\renewcommand{\shorttitle}{Graph-Based Soft-Label Imitation Learning for Power Grid Control}

\hypersetup{
pdftitle={Learning Topology Actions for Power Grid Control: A Graph-Based Soft-Label Imitation Learning Approach},
pdfsubject={q-bio.NC, q-bio.QM},
pdfauthor={David S.~Hippocampus, Elias D.~Striatum},
pdfkeywords={First keyword, Second keyword, More},
}

\begin{document}
\maketitle

\begin{abstract}
The rising proportion of renewable energy in the electricity mix introduces significant operational challenges for power grid operators. Effective power grid management demands adaptive decision-making strategies capable of handling dynamic conditions. With the increase in complexity, more and more Deep Learning (DL) approaches have been proposed to find suitable grid topologies for congestion management. In this work, we contribute to this research by introducing a novel Imitation Learning (IL) approach that leverages soft labels derived from simulated topological action outcomes, thereby capturing multiple viable actions per state. Unlike traditional IL methods that rely on hard labels to enforce a single optimal action, our method constructs soft labels that capture the effectiveness of actions that prove suitable in resolving grid congestion. To further enhance decision-making, we integrate Graph Neural Networks (GNNs) to encode the structural properties of power grids, ensuring that the topology-aware representations contribute to better agent performance. Our approach significantly outperforms its hard-label counterparts as well as state-of-the-art Deep Reinforcement Learning (DRL) baseline agents. Most notably, it achieves a 17\% better performance compared to the greedy expert agent from which the imitation targets were derived.

\end{abstract}

\section{Introduction}
In recent years, \gls{rl} and \gls{il} have emerged as powerful approaches for sequential decision-making in complex environments, including power grid management. In this context, agents must make rapid and informed topological adjustments to maintain grid stability under dynamic conditions. Recent advances in power grid control have demonstrated the effectiveness of \gls{rl}-based agents, particularly when they are pre-trained using \gls{il} \cite{binbinchen,lehna2023managing}. Prior work has applied \gls{il} for topology control using standard feed-forward neural networks with subsequent \gls{rl} fine-tuning to improve decision-making policies \cite{lehna2024hugo,lehna2023managing,binbinchen}. Additionally, \glspl{gnn} have become popular as a structured way to encode power grid topology, enabling improved action representation and decision-making \cite{grlsurvey2024,taha_learning_2022,xu_active_2022,sar_multi_agent_2023,yoon2021winning,zhao2022}. 

However, the existing \gls{il} methods often fail to capture the inherent uncertainty in the solution space and typically learn to mimic a single expert action per state, disregarding the fact that there are often multiple effective interventions that can ease congestion. This restrictive view can undermine policy robustness and adaptability, leading to rigid policies that struggle with generalization. 

Grid2Op \cite{grid2op} provides a widely used framework for developing and evaluating \gls{rl}-based grid control methods, particularly for topology optimization tasks such as substation reconfigurations \cite{rl2grid}. In addition, \gls{rl} agents have demonstrated strong performance in \gls{l2rpn} challenges \cite{marot2020learning,marot2020l2rpn,marot2022learning,serre2022reinforcement}, where the goal is to maintain grid operability under uncertainty and disturbances. In our experiments, we leverage the WCCI 2022 environment implemented in Grid2Op\footnote{Grid2Op: \url{https://grid2op.readthedocs.io/en/latest/} (last accessed 12/03/2025).}, thus allowing the benchmarking of our methods.

\subsection{Main Contributions}
To address the limitations of current approaches and enable more robust and adaptable policy learning for power grid control, we propose a soft-label imitation learning approach. Soft-label \gls{il} retains  and exploits information about multiple effective actions for each grid state through a richer supervisory signal. This rich supervision guides the policy toward greater robustness and adaptability, reflecting the operational reality that power grid congestion can be resolved in more than one single way. Our approach thereby avoids overfitting to potentially sub-optimal expert decisions, reduces label noise, and guides the agent in learning a generalized policies that also are applicable for previously unseen grid states. Furthermore, soft labels enable us to naturally produce a ranking of candidate actions, which is especially valuable in power grid control, where the choice of multiple viable interventions can account for operational preferences, N-1 contingencies, or robustness criteria. This combination -- retaining multiple desirable options alongside their respective confidence scores -- ultimately results in a more reliable, adaptable, and realistic control policy.

Additionally, we leverage \glspl{gnn} to account for the structural properties of power grids, reflecting their physical topology and power flow relationships. GNNs enable the policy to learn contextually rich representations for each grid component, which further improves decision-making. Our contributions can be summarized as follows:
\begin{enumerate}
    \item Development of a novel soft-label approach for \gls{il} in power grid control, incorporating multiple viable actions into the learning signal.
    \item The integration of \glspl{gnn} to effectively leverage the inherent graph structure of power grids and enhance decision-making.
    \item A demonstration that our method outperforms two state-of-the-art \gls{rl} approaches and particularly the greedy expert itself by utilizing soft action labels.

\end{enumerate}

\section{Related Work}
The idea of congestion management through topology optimization has witnessed a surge in research interest, in part due to the \gls{l2rpn} challenges by the french \gls{tso} RTE \cite{marot2020learning,marot2020l2rpn}. In many cases the proposed solutions consist of a model-free \gls{drl} algorithm that is restricted by rule-based or heuristic components \cite{binbinchen,lehna2024hugo,lehna2023managing,chauhan2022powrl,zhou2021action}. Most of these \gls{drl} approaches are built with standard \gls{fnn}, however, \cite{grlsurvey2024} find in their survey that an increasing number of researchers use \glspl{gnn} to incorporate the graphical nature of the power grid. As the number of topology actions increases drastically with grid size, there have further been different approaches to tackle the large action spaces. Some researchers propose a hierarchical agent strategy \cite{manczak2023hierarchical,yoon2021winning},  or multi-agent approaches \cite{sar_multi_agent_2023,demol2025} to split the decision making process in smaller sub-tasks. Alternatively, \cite{dorfer2022power} propose a \gls{mcts} to plan multiple steps ahead.

Moreover, \gls{il} has been explored for power grid control, motivated by the potential to accelerate computation through the imitation of rule-based and other expert agents. While there have been some application of \gls{il} by \cite{binbinchen,lehna2023managing} and \cite{lehna2024hugo} to pre-train a feed-forward network on a greedy agent, they only used the models to jumpstart the \gls{drl} training process but didn't utilize the \gls{il} model as an agent for topolgy control directly. A further \gls{il} approach in this regard has been studied by \cite{dejong2024} and \cite{dejong2025}. In the first paper, \cite{dejong2024} analyze both a greedy and a N-1 rule-based agent on the Grid2Op IEEE 14 environment and then use the experience of the agents to train a \gls{il} model. Several types of hybrid agents were constructed, which combined \gls{il} and simulation functionality. The hybrid agents showed similar performance with almost 100\% completion of the scenarios, while reducing the inference duration of the agent. Even more interesting, \cite{dejong2024} found that there occurs in some cases a confusion of the actions by the \gls{il} model, as some actions are not clearly distinguishable in some scenarios. In the second paper, \cite{dejong2025} follow up on their \gls{il} framework and focus on applying (node-level) \gls{gnn} prediction of the grid topology. They identify the busbar information asymmetry problem, where nodes on the same substation but different busbars remain unconnected in traditional graph representations, hindering GNN performance. They propose a heterogeneous GNN to address this by modeling inter-busbar connections, outperforming homogeneous \glspl{gnn} and \glspl{fnn} in accuracy and out-of-distribution generalization. Existing \gls{il} methods inherit expert biases by relying on deterministic policies that overlook diverse viable actions for overload mitigation, we highlight the need to capture all effective actions instead and address it through a richer representation of the labels.

\section{Power Grid Setup}
As mentioned earlier, we follow the previous researchers and use the Grid2Op environment, as it is the current benchmark for transmission grids \cite{grlsurvey2024} and allows the comparison with other approaches. Grid2Op is an open-source simulation platform designed for power grid operation research, particularly in the context of \gls{drl} and other \gls{dl} control strategies. 
Since Grid2op was designed with \gls{rl} in mind, we utilize the same terminology, though we do not apply \gls{rl} in this work.

\subsection{Environment} \label{ssec:env}
In this work, we utilize the L2RPN WCCI 2022 environment, which models the IEEE 118-bus transmission system with an expected 2050 electricity mix. As a result, the simulated fossil fuel generation accounts for less than 3\%, and renewable energy sources are significantly increased \cite{serre2022reinforcement}. 
\begin{figure}[b!]  
    \centering
    \includegraphics[width=\linewidth]{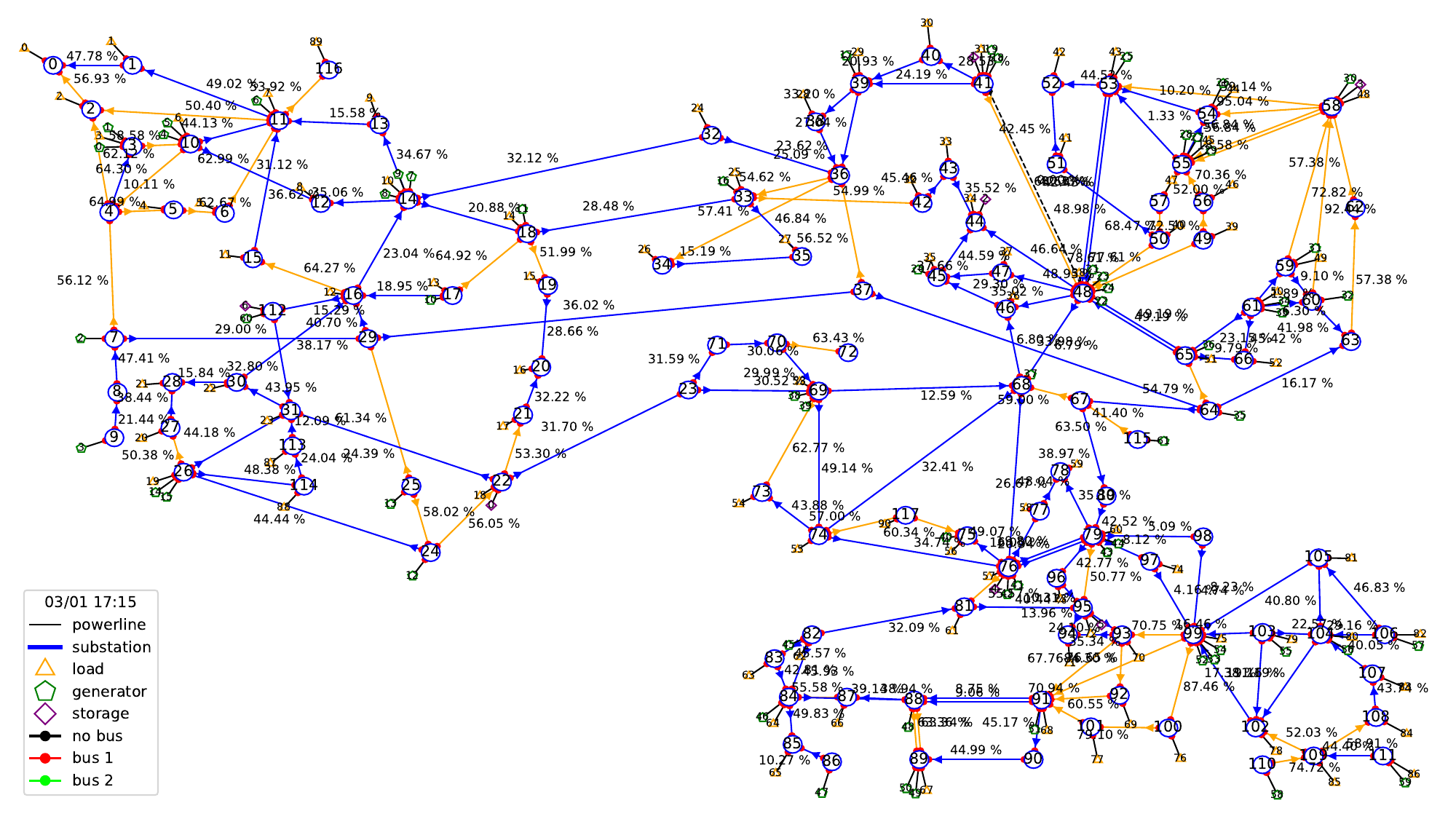} 
    \caption{Overview of the WCCI 2022 L2RPN environment, generated using Grid2Op’s native visualization tools.}
    \label{fig:grid2022} 
\end{figure}
At its core, the power grid can be represented as a graph where substations are nodes that are connected via transmission lines. Substations serve as connection points for grid components, including generators, loads, and power lines. This can be seen in Fig. \ref{fig:grid2022}, where we visualize the WCCI 2022 environment. This specific IEEE 118 transmission grid consists of 118 substations, 91 loads, 62 generators, and 7 battery storage units, all interconnected by 186 transmission lines. The observation space $S$ of the contains 4,295 features such as active and reactive power flows, voltage magnitudes and angles, generator and storage injections, load demands, planned maintenance schedules, cooldown periods, and topology configurations. Among these, the most critical variable for this work is the line loading capacity, denoted as $\rho_l$ for each line $l \in L$, with the maximum capacity across all lines given by $\rho_{max} = \max_{l}(\rho_l)$. Furthermore, a double busbar system is implemented. There, each substation consists of two busbars, with each component connected to either one of the busbars. However, power can only flow between elements connected to the same busbar within a substation. Thus, through the reassignment of grid components to a different busbar, one can substantially alter the power flow. For this reason, the topology optimization on a substation can provide a cost-effective and fast solution to tackle congestion issues in the transmission grid. Another feature of Grid2Op is the simulation function, \verb|obs.simulate()|, which is essential for assessing the impact of proposed actions. It forecasts the grid’s next state based on realistic generation and consumption data and computes the power flow. 

Power flows can be simulated under the N-0 case, which assumes normal grid operation without any line outages. In contrast, the N-1 case involves simulating the failure of a single transmission line and calculating the resulting power flows, particularly focusing on the maximum line loadings to assess grid resilience. In the Grid2Op framework, a full N-1 contingency analysis is not conducted exhaustively. Instead, a softened version is used where an adversarial agent randomly disconnects one line from a predefined subset of lines deemed critical \cite{rl2grid}. This simulates unexpected failures while maintaining computational feasibility and allowing learning-based agents to develop robust control strategies.

Grid2Op simulates power grid operations in 5‑minute intervals, modeling fluctuations in demand, generation adjustments, and potential failures. Its goal is to maintain stability and prevent cascading failures due to line overloads using synthetic scenarios (\textit{chronics}) based on historical data. An episode in Grid2Op can end in two ways: successful completion or early termination due to grid failure. A successful episode occurs when an agent manages the grid throughout all 2016 time steps (equivalent to one week). In contrast, early termination -- often resulting in a blackout -- happens when grid stability is compromised. A common cause is cascading failures triggered by the rule that disconnects a transmission line if its load remains above $100\%$ for three consecutive time steps. Additionally, an episode ends immediately, in case a generator or load is disconnected, islanding occurs, or if the power flow solver fails to converge.
\begin{figure}[b!]
    \centering
    \includegraphics[trim={0.0cm 5.0cm 2.0cm 3.42cm},
    clip, width=1.0\linewidth,page=3]{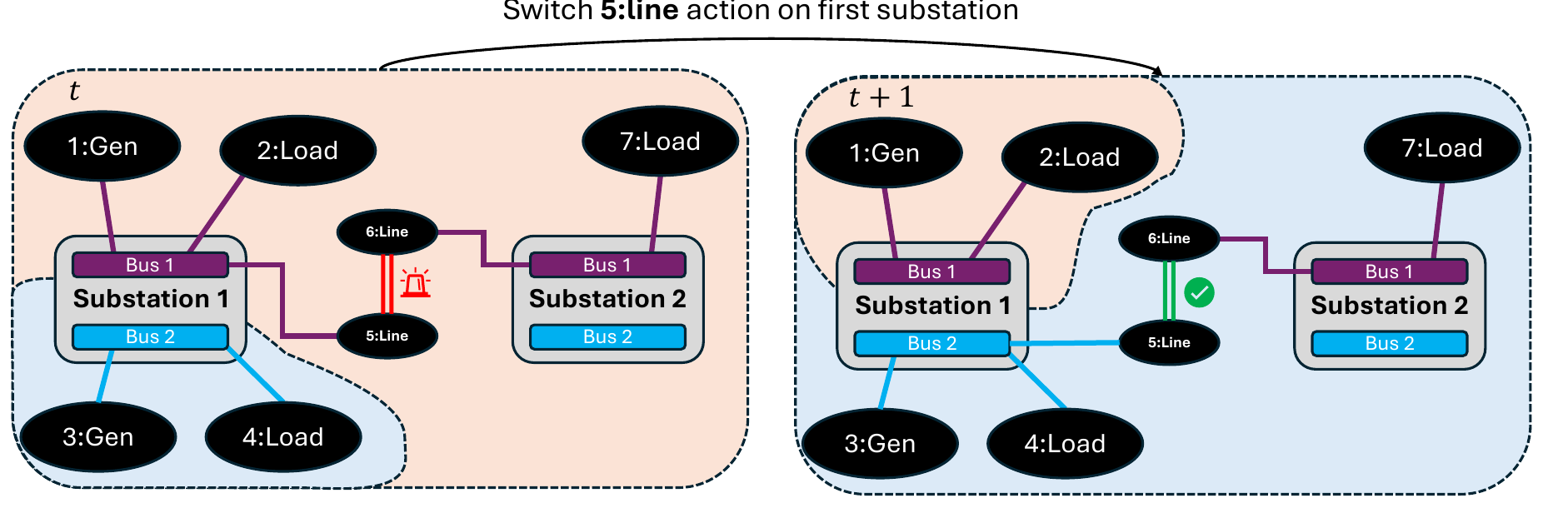} 
    \caption{Example of a substation reconfiguration action. Two substations are visualized, each with two busbars. Generators, loads, and line ends are represented as nodes that are interconnected via a busbar of a substation. From time step $t$ to $t+1$, one end of line 5 is switched to the other busbar, altering the connection between the elements and mitigating the line overload.}
    \label{fig:action} 
\end{figure}
\subsection{Action Space}
The action space is divided into four different action types. The first action type includes line disconnection and reconnection, while the second type includes substation reconfiguration, which we interchangeably refer to as topology actions. The third action type includes generator redispatch as well as curtailment of renewable energies, while the fourth type relates to battery storage operations. In this work however, we only consider topology actions as they are the most cost-efficient action type. One simplified example substation reconfiguration action is shown in Fig. \ref{fig:action}, which switches the line end 5 from the first to the second busbar. This completely changes the connection between the two substations, altering the power flow and mitigating the line overload. These discrete actions scale combinatorially with the number of substation elements, creating a large action space. Grid2Op restricts certain actions to maintain operational feasibility, ensuring that agents cannot take unrealistic or physically impossible steps. Nevertheless, the large action space still presents a significant challenge for control algorithms, hence an action space reduction is usually applied. Note that the line disconnection and reconnection also influences the busbar configuration, as non-connected elements can not be switched by topology actions. The resulting implications are later discussed in Sec. \ref{ssec:fix}, as they influence the feasibility of topology actions. Nevertheless, all compared agents, including our own, do not explicitly learn line disconnection and reconnection, but instead automatically reconnect lines whenever possible to restore grid stability.
\section{Methodology}
We present a systematic approach to addressing grid overloads by combining simulation-based decision-making with advanced learning techniques. We begin by detailing the \gls{greedy}, a reactive agent that selects the best corrective actions based solely on current grid conditions via load flow simulations. Then, we introduce the generation of soft labels, a strategy that leverages the full spectrum of simulated outcomes to create a richer supervisory signal. Additionally, we explain how the inherent graph structure of power grids is leveraged using \glspl{gnn} to capture spatial dependencies. Lastly, the agent's architecture and practical enhancements, such as topology reversion and improved action feasibility is discussed.
\subsection{Greedy Expert}
The \gls{greedy} performs an N-0 load flow simulation for all possible actions for one timestep and selects the one that has the lowest maximum line loading. For this, we utilize Grid2Op's \verb|env.simulate()| method, which perform a power flow calculation based on forecasts of generation and loads. Specifically, the \gls{greedy} is activated if any line loading exceeds 90\%. In that case it simulates all actions and selects the action that provides the lowest maximum line loading $\rho_{max}$.
This agent requires a lot of costly power flow simulations and is highly reactive, making decisions based solely on the current state of the grid without considering long-term implications or alternative strategies. Furthermore, it only considers the maximum line loading value and thus other lines with higher load are not regarded. Consequently, it is missing exploration and does not consider the holistic effect of actions.

\subsection{Soft Labels}
We begin the generation of soft labels by running the \gls{greedy} agent through our environment. The details of the complete procedure for generating soft labels along with applying the greedy optimal action are described in Algorithm \ref{algo:data_collection}.
However, instead of a simple greedy iteration through all actions and simulating the $\rho_{max}$ for each action (line 4\&5), we also compute an effectiveness score $e_a$ based on the inverse maximum line loading $1- \rho_{max}$ (line 5).\footnote{Note that we do not collect an effectiveness score if a grid failure is imminent, i.e., no action is able to resolve the congestion in the next step.} With this effectiveness score, we generate our soft labels that compare the impact of the action to all other actions. These soft labels are created by applying a temperature softmax function to the effectiveness score $e_a$ (line 8). Note that we used a temperature parameter $\tau = 0.01$ via preliminary tests over several values to sharpen the softmax so that highly effective actions get substantial probability mass without excessive skew; a full sensitivity analysis could further validate this choice.


The remaining algorithm is then simply the greedy selection from the original \gls{greedy} (line 11).
\begin{algorithm}
    \caption{Soft Label Generation}
    \label{algo:data_collection}
    \begin{algorithmic}[1]
        \Require Environment \texttt{env}, Set of actions $\mathcal{A}$, temperature parameter $\tau$

        \State Initialize dataset $\mathcal{D} \gets \emptyset$ and environment \texttt{env}
        \For{each observation $s \in S$} \Comment{Iterate through environment and receive $s$}
            \For{each action $a \in \mathcal{A}$} \Comment{Simulate maximum Line Loads}
                \State Run \texttt{env.simulate($a$)} to get $\rho_{max}(s, a)$
                \State Compute effectiveness score: $e_a = 1 - \rho_{max}(s, a)$
            \EndFor
            \For{each action $a \in \mathcal{A}$} \Comment{Compute soft labels}
                \State $\Psi(a \mid s) = \frac{\exp(e_a / \tau)}{\sum_{a' \in \mathcal{A}} \exp(e_{a'} / \tau)}$
            \EndFor
            \State Store $(s, \Psi(a \mid s))$ in dataset $\mathcal{D}$
            \State Apply greedy optimal action $a^* = \arg\min\limits_{a' \in \mathcal{A}} \rho_{max}(s, a') $
 using \texttt{env.step($a^*$)} 
        \EndFor
        \State \Return $\mathcal{D}$ 
    \end{algorithmic}
\end{algorithm}
Ultimately, this approach enables the agent to learn from all the simulations of possible actions that are otherwise discarded. Soft labels provide a relative measurement. A significantly higher value for a particular action indicates that this action is distinctly more effective in reducing line load compared to its counterparts and therefore, the model should be more confident in predicting this action. Conversely, if the effectiveness ratio is more evenly spread across several actions, it suggests that these actions have similar effects. Instead of rigidly following the best action, the agent is exposed to a richer supervisory signal. 

\subsection{Utilizing the inherent graph structure of power grids}
Power grids are inherently graph-structured systems, where substations, generators, loads, and transmission lines form interconnected nodes and edges. To leverage this structure, we transform the observations into a graph representation. This graph-based encoding allows \glspl{gnn} to capture spatial dependencies and propagate congestion-reduction strategies across the grid.

\subsubsection{Graph Construction}
We treat each grid component -- loads, generators, each end of a transmission lines, and storage --  as an individual node. For every node, we aggregate features that capture the state of the grid component. To construct the graph, we first extract a range of features from the observation. These features include operational parameters like cooldown values, power injections, voltage measurements, and maintenance information. Each row in the graph's feature matrix corresponds to a grid asset, and each column captures a specific attribute, e.g., the power consumption of a load or the generation capacity of a generator. Missing values for features that do not apply to a particular component type are filled with zeros.

The edges in the graph are determined based on the physical connectivity of the grid. Specifically, nodes are connected according to the grid's topology, where edges represent electrical connections between components within the same substation or via transmission lines linking different substations. Importantly, transmission line features such as power flow, voltage, and loading are encoded directly into the nodes representing the respective line ends. This approach allows for a uniform node-based feature representation, ensuring that all relevant grid information is captured at the node level while maintaining a simple and efficient graph structure. 
\subsubsection{GNN architecture}
Our GNN architecture employs \gls{gat} \cite{gat_paper} to model relationships between grid components, using attention mechanisms to weight the influence of neighboring nodes and thereby prioritizing critical connections, such as heavily loaded lines. Each observation is first transformed into a graph structure and processed through four GAT layers, which progressively refine the node representations. A global max pooling operation then aggregates these node-level features into a representation for graph-level prediction. This pooled output is subsequently passed through three feed-forward layers and an output layer whose dimensionality matches the size of the action space. The architecture search was conducted using Optuna’s \gls{tpe} algorithm \cite{tpe}, ensuring that the model's hyperparameters were optimally tuned for the task.
Finally, the model is trained using Kullback-Leibler Divergence (\verb|KLDivLoss|) to minimize the discrepancy between predicted output and soft labels derived according to Algorithm \ref{algo:data_collection}.

\subsection{Agent}
We construct the \textit{SoftGNN}$_{90\%}$ agent using the \gls{gnn} with specific mechanics to ensure adequate performance. First, we iterate through the environment and activate the agent only in case of an emergency, i.e., when the max line load $\rho_{max}$ exceeds the threshold of 90\%. In case the grid is stable ($\rho_{max}<90\%$), we either execute a \textit{DoNothing} action or revert the topology, as described below. 
Otherwise, we use the model’s predictions to the current observation and sort the actions in descending order according to the model's output. The agent then iterates through the sorted list, validating each candidate action for feasibility and simulate its impact on the grid using the \verb|env.simulate()| method to compute the post-action maximum line loading $\rho_{max}$. The first action that reduces $\rho_{max}$ below the predefined threshold of 90\% is executed. If no such action exists, the agent defaults to the \textit{DoNothing} action. This process ensures that if an action is selected, it mitigates the grid congestion while adhering to operational constraints. We describe the ad-hoc enhancements as follows: 

\subsubsection{Topology reversion}
\label{ssec:revert}
Reverting back to the base topology of the power grid in which all busbar couplers are closed, i.e., no substation is split, has been shown to enhance the performance of agents \cite{lehna2024hugo,lehna2023managing}. This is due to the fact that the base topology performs well for stable time steps such as during nights. Therefore, all agents discussed in this paper check for topology reversion when the activation threshold has not been reached. Whenever safely possible, i.e., when reversion doesn't cause $\rho_{max}$ to increase beyond the threshold of 80\%, the topology reversion is applied. The threshold value is adopted based on established literature \cite{lehna2024hugo}, which has demonstrated that this value yields optimal performance.

\subsubsection{Enhancing Topology Action Feasibility} \label{ssec:fix}
In the context of applying substation reconfiguration actions in Grid2Op, we identified a critical limitation in the default behavior of the topology actions using the \verb|set-bus| method: when an action is applied, it attempts to set busbar assignments for all elements, including those associated with currently disconnected lines. This behavior is problematic because it invalidates actions that are subsequently rejected by the environment. Particularly, all actions of both substations adjacent to the disconnected line are rendered invalid. This can be very critical when the disconnected line is adjacent to a large substation that accounts for a significant proportion of the action space.
To address this issue, we introduce a pre-processing step that selectively removes bus assignments for all disconnected lines before applying an action. By doing so, the action retains valid bus assignments for connected components while ensuring that no invalid modifications are attempted for disconnected lines and hence significantly improving the feasibility of topology actions. 

\subsubsection{N-1 load flows at inference time}
 \label{ssec:n1}
Lastly, to enhance robustness against potential line failures, we propose an extension agent \textit{SoftGNN}$_{90\%}\, N-1$ that incorporates N-1 security criteria into its action selection. During inference, we first filter the top 10 actions from the sorted \gls{gnn} output. For each candidate action, the agent simulates its impact under both N-0 (no failures) and N-1 (single-line failure) scenarios. It prioritizes actions that minimize this worst-case metric and selects the action with the lowest N-1 $\rho_{max}$. However, if all N-1 simulations result in overloads, i.e., $\rho_{max}\geq 100\%$, the agent falls back to the N-0 criterion, selecting the action that minimizes the baseline $\rho_{max}$. Consistent with the Grid2op framework and \cite{lehna2024hugo}, we do not perform a full N-1 contingency analysis across all lines, but instead restrict simulations to the predefined subset of lines that can be attacked by the adversarial agent. Furthermore, we exclude line 93 from the N-1 analysis in accordance with the findings of \cite{lehna2024fault}, which show that disconnecting this line inevitably triggers a cascading failure within three time steps, regardless of the remedial action taken. Including such a pathological case would disproportionately distort the evaluation of otherwise effective actions.

\section{Experiments}
\subsection{Experimental Setup}
As outlined in Sec. \ref{ssec:env}, our study utilizes the WCCI 2022 L2RPN framework, visualized in Fig. \ref{fig:grid2022}. We train our agents on the publicly accessible environment data. All agents are trained for a maximum of $800$ epochs, however, early stopping is applied based on validation performance to prevent overfitting. The early stopping criterion monitors the validation loss, and training halts if no improvement is observed for $20$ consecutive epochs. The model with the best validation performance is selected for evaluation. Each agent is trained using the Adam optimizer, with learning rate adjustments managed by a learning rate scheduler that reduces the learning rate by a factor of $0.9$ if no improvement in the validation loss is observed for $10$ consecutive epochs. The batch size was fixed to $256$ while all other hyperparameters were determined using optuna. Table \ref{tab:gat_hyperparameters} shows the selected hyperparameters for the \gls{gat} model of all \textit{SoftGNN} agents as well as the respective search ranges. The \gls{gat} layers apply ELU as their activation function, while the subsequent linear layers apply  ReLU. Hyperparameter search was conducted in a distributed setup on a computing cluster featuring 8 NVIDIA A100 GPUs, and training of the final model required approximately 8 hours on a single GPU.
\begin{table}[t!]
\caption{Model architecture of the soft-label graph neural network used all \textit{SoftGNN} agents. The architecture search was conducted using Optuna's Tree-structured Parzen Estimator (TPE) \cite{tpe}. Additionally, the search range of the dropout parameter for all layers is $[0 , 0.5]$.}
\centering
\renewcommand{\arraystretch}{1.2} 
\setlength{\tabcolsep}{4pt}   
\resizebox{\textwidth}{!}{
\begin{tabular}{lll}
\toprule
\textbf{Component} & \textbf{Hyperparameters (Selected)} & \textbf{Search Range} \\
\midrule
\multicolumn{3}{c}{\textbf{GAT Layers (input, output, heads, dropout)}} \\
\midrule
Layer 0 & GATConv(27, 16, 1, 0.177) & dim: \{16, 32, 64, 128\}, heads: \{1,2,4,8\}\\
Layer 1 & GATConv(16, 32, 4, 0.139) & dim: \{16, 32, 64, 128\}, heads: \{1,2,4,8\} \\
Layer 2 & GATConv(128, 64, 2, 0.174) & dim: \{16, 32, 64, 128\}, heads: \{1,2,4,8\}\\
Layer 3 & GATConv(128, 128, 4, 0.096) & dim: \{16, 32, 64, 128\}, heads: \{1,2,4,8\} \\
Num. of Layers & 4 & \{2, 3, 4, 5, 6\}\\
Pooling & Global Max Pooling & \{"max", "mean", "add"\} \\
\midrule
\multicolumn{3}{c}{\textbf{Linear Layers (input, output, dropout)}} \\
\midrule
Layer 0 & Linear(128, 1024, 0.496) & dim: \{128, 256, 512, 1024\}\\
Layer 1 & Linear(1024, 1024, 0.489) & dim: \{128, 256, 512, 1024\} \\
Layer 2 & Linear(1024, 2030, 0.0) & — \\
Num. of Layers & 3 & \{2, 3, 4\}\\

\midrule
\multicolumn{3}{c}{\textbf{Training}} \\
\midrule
Learning Rate & $4.14 \times 10^{-3}$ & [$10^{-5}$–$10^{-2}$] \\
Weight Decay & $8.48 \times 10^{-6}$ & [$10^{-6}$–$10^{-3}$] \\
\bottomrule
\end{tabular}}
\label{tab:gat_hyperparameters}
\end{table}

We evaluate our agents using the test environment of the 2022 challenge \cite{serre2022reinforcement} provided by RTE France. The test environment comprises 52 scenarios, each spanning 2016 time steps. We follow the approach of \cite{lehna2024hugo} by employing 20 randomized master seeds to ensure statistical robustness and address variability across scenarios influenced by environmental seed differences. For comparability, we use the same master seeds. We further use the same action space of \cite{lehna2024hugo}, which consists of $2000$ actions from the \gls{l2rpn} 2022 challenge winner \cite{dorfer2022power}. Moreover, $30$ expert actions selected by RTE were added, resulting in a total of $2030$ actions.

The dataset, code, and trained models will be made publicly available in a dedicated GitHub repository\footnote{\url{https://github.com/AI4REALNET/soft\_label\_gnn}}, as well as to the \textit{CurriculumAgent}\footnote{\url{https://github.com/FraunhoferIEE/curriculumagent}} repository for compatibility with state-of-the-art \gls{rl} approaches.

\subsubsection{Ablation Study}
Our ablation study evaluates the impact of soft labels and \gls{gnn} models on agent performance by comparing four core variants, which allows for a granular comparison. The variants include two hard-label approaches—an FNN model and a GNN model—termed \textit{HardFNN} and \textit{HardGNN}, respectively, as well as two soft-label approaches—\textit{SoftFNN} and \textit{SoftGNN}. Additionally, we analyzed the \textit{SoftGNN}$_{90\%}\, N-1$ variant from Sec. \ref{ssec:n1} to assess the synergy between soft labels and N-1 safety-aware action selection. The agents were trained on the same data generated using Algorithm \ref{algo:data_collection}, with identical action spaces and the pre-processing fix for invalid bus assignments described in Sec. \ref{ssec:fix}. We further compare the performance of these models to four benchmark agents. First the \textit{DoNothing} baseline, second the expert greedy agent \gls{greedy} with the pre-processing fix. Moreover, we re-evaluate two state-of-the-art agents from literature (\textit{Senior}$_{95\%}$ \cite{lehna2024hugo} and \textit{TopoAgent}$_{85-95\%}$ \cite{lehna2024hugo}) with the pre-processing fix in order to isolate the effects of the fix. We dub these agents \textit{SeniorFix}$_{95\%}$ and \textit{TopoAgentFix}$_{85-95\%}$, respectively.

The \textit{Senior}$_{95\%}$ is a sophisticated DRL agent that performs topology actions when $\rho_{max}$ exceeds a $0.95$ threshold, ensuring safe and reliable intervention during extreme conditions, while the superior \textit{TopoAgent}$_{85-95\%}$ activates under moderate instability and additionally employs a greedy search over pre-identified robust Target Topologies to sequentially combine actions and guide the grid toward a more stable configuration.

\subsubsection{Metrics}
Performance metrics include the \gls{l2rpn} score (mean, median, quartiles) from the \gls{l2rpn} 2022 challenge \cite{serre2022reinforcement}, a composite score that assesses the agent's ability to keep the power grid operational while minimizing operational costs. The score is computed by first calculating the total operational cost for each scenario -- this includes energy losses, redispatch, curtailments, storage operations, and penalties for blackouts -- and then applying a linear transformation to aid interpretability. It is calibrated by assigning the  \textit{DoNothing} baseline a score of $0$. Agents performing worse than this baseline can receive scores as low as $-100$, while those that survive longer earn positive scores. The completion of every episode results in a score of $80$ and for a $100$ the agent must also minimize both energy loss and operational costs. Moreover, we measure the survival time with the median survival time and the \gls{mstcm}. The latter is less influenced by outlier performance since it averages over the chronics first \cite{lehna2024hugo}.
\begin{figure}[h]  
    \centering
    \includegraphics[width=\linewidth]{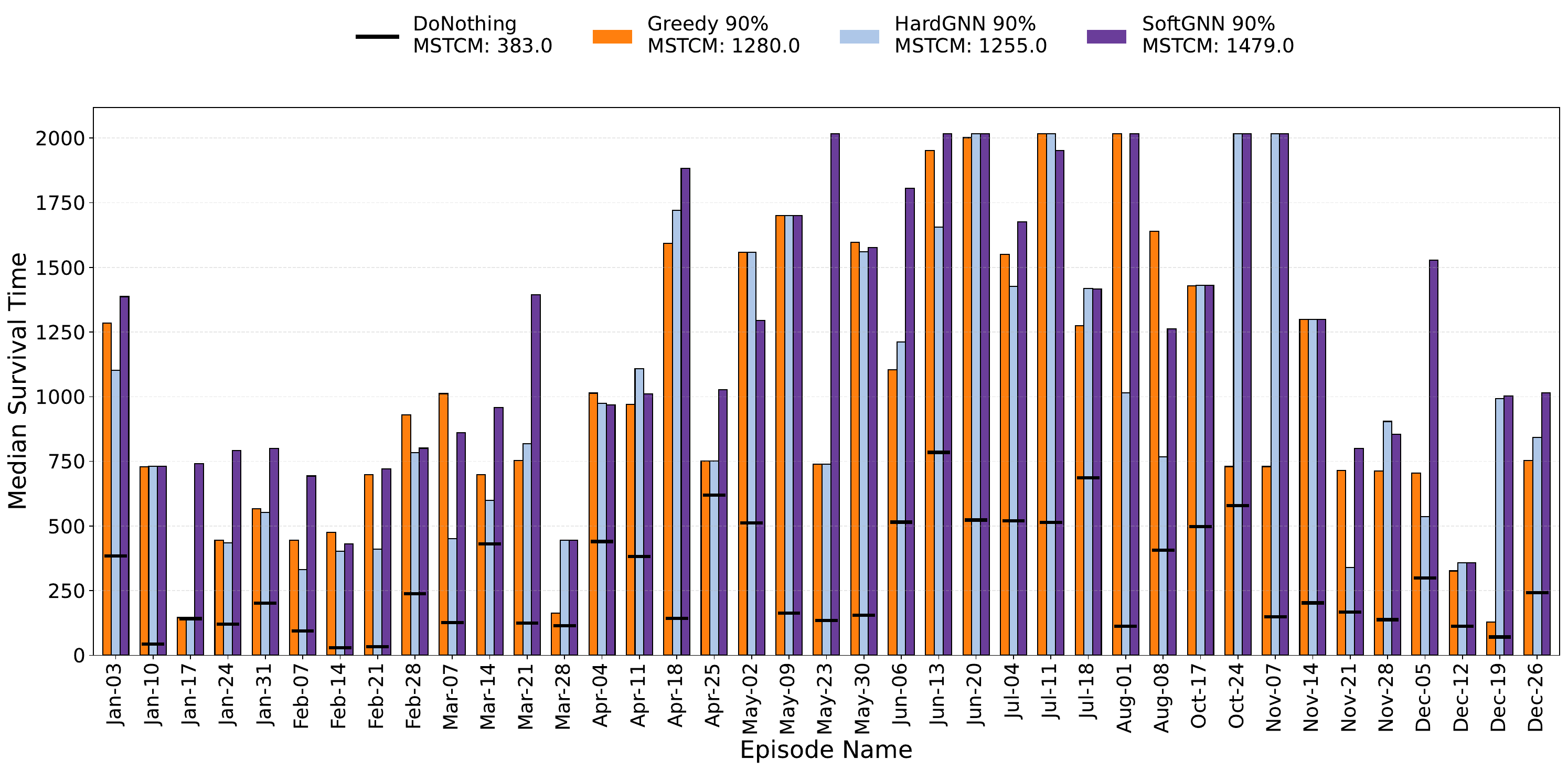} 
    \caption{Comparison of agent Median Survival Times across all test scenarios, calculated over 20 random seeds. The \gls{mstcm} is shown above the figure for reference. Chronics where all non-baseline agents survived (median survival time of $2017$ steps) were excluded (13 in total) for clarity.}
    \label{fig:survival} 
\end{figure}
\subsection{Results}
The experimental evaluation demonstrates the efficacy of our soft-label imitation learning approach combined with \glspl{gnn}. We summarize the results in Table \ref{tab:agent_performance} and visualize in Figure \ref{fig:survival} the median survival time of each chronic across the 20 seeds. 
As expected, the \textit{DoNothing} baseline achieved a score of $0$ and median survival time of 229 steps, while the \gls{greedy} agent improved performance with a mean L2RPN score of $37.91$ and a median survival time of $1014$ steps. 

Re-evaluating previous state-of-the-art agents with the action feasibility fix (Sec. \ref{ssec:fix}) yielded measurable gains. The \textit{SeniorFix}$_{95\%}$ agent outperformed the original \textit{Senior}$_{95\%}$, increasing its median \gls{l2rpn} score from 37.13 to 39.40 and \gls{mstcm} from $1160$ to $1468$. Similarly, \textit{TopoAgentFix}$_{85-95\%}$ achieved higher performance across all metrics, though to a lesser extent. This clearly shows the effect of our preprocessing-fix.

With respect to the ablation study, both hard-label approaches perform similarly to the \gls{greedy} agent with a small advantage of the \textit{HardGNN}$_{90\%}$. This highlights the struggles to overcome imperfect teacher agents. In contrast, the soft-label models significantly surpass their hard-label counterparts. Among the soft-label agents, the \gls{gnn} agent outperforms the \gls{fnn} variant, indicating a synergy between the enhanced \gls{gnn} feature representation and soft label learning. The \textit{SoftGNN}$_{90\%}$ agent improved the \gls{l2rpn} score by nearly 15\% compared to its hard-label counterpart \textit{HardGNN}$_{90\%}$. Particularly for very challenging runs, the \textit{SoftGNN}$_{90\%}$ agent manages to outperform the hard-label variant and the expert and survives significantly longer. Similarly, for the \gls{fnn} variants, the soft-label agent improved the score by 8\%.

Finally, our  \textit{SoftGNN} agents are able to outperform the state-of-the-art agents \textit{Senior}$_{95\%}$ and \textit{TopoAgent}$_{85-95\%}$ that employ more sophisticated simulation strategies and specifically optimize for long-term performance through \gls{rl}. The \textit{SoftGNN}$_{90\%}\, N-1$, incorporating N-1 security criteria during inference, achieved the highest overall performance with a mean \gls{l2rpn} score of $44.43$ and median survival time of $1299$ steps. Notably, its \gls{mstcm} of $1566$ surpassed even the \textit{TopoAgentFix}$_{85-95\%}$ by $72$ time steps, demonstrating the synergy between soft-label learning and safety-aware action selection. 
\begin{table}[t!]
    \caption{Overview of the aggregated agent performance. The table provides the L2RPN score metric statistics as well as the median survival time and \gls{mstcm}.}
    \centering
    \renewcommand{\arraystretch}{1.2} 
    \setlength{\tabcolsep}{3pt}       
    \begin{tabular}{lccccc|cc}
        \toprule
        \multirow{2}{*}{\textbf{Agent}} & \multicolumn{5}{c}{L2RPN Score} & \multicolumn{2}{c}{Survival Time} \\
        \cmidrule(lr){2-6} \cmidrule(lr){7-8}
         & $\boldsymbol{\bar{x}}$ & $\boldsymbol{\sigma}$ & $\boldsymbol{\tilde{x}}$ & $\boldsymbol{Q_1}$ & $\boldsymbol{Q_3}$ & $\boldsymbol{\tilde{x}}$ & \textbf{MSTCM}\\
        \midrule
        \textit{DoNothing} & 00.00 & 0.00 & 00.00 & 00.00 & 00.00 & 229 & 383 \\
        \midrule
        $Greedy_{90\%}$ & 37.91 & 3.89 & 37.07 & 35.62 & 40.78 & 1014 & 1280\\
        \midrule
        \textit{Senior}$_{95\%}$ \cite{lehna2024hugo} & 37.13 & 4.49 & 37.21 & 33.48 & 39.84 & 988 & 1160 \\
        \textit{SeniorFix}$_{95\%}$ & 39.40 & \textbf{2.98} & 39.50 & 37.14 & 41.25 & 1026 & 1468 \\
        \midrule
        \textit{TopoAgent}$_{85-95\%}$ \cite{lehna2024hugo} & 41.26 & 3.01 & 40.41 & 39.41 & 43.69 & 1232 & 1436 \\
        \textit{TopoAgentFix}$_{85-95\%}$ & 41.81 & 3.11 & 42.17 & 40.33 & 44.00 & 1263 & 1494 \\
        \midrule
        \textit{HardFNN}$_{90\%}$ & 37.54 & 3.87 & 37.08 & 35.06 & 40.22 & 1020 & 1114\\
        \textit{SoftFNN}$_{90\%}$ & 40.73 & 4.16 & 40.54 & 36.60 & 43.64 & 1113 & 1316\\
        \midrule
        \textit{HardGNN}$_{90\%}$ & 38.28 & 3.58 & 37.95 & 35.85 & 40.08 & 1048 & 1255\\
        \textit{SoftGNN}$_{90\%}$ & 43.84 & 3.60 & \textbf{43.96} & 41.40& 46.09 & 1293 & 1479\\
        \textit{SoftGNN}$_{90\%}\, N-1$ & \textbf{44.43} & 3.27 & 43.49 & \textbf{42.33} & \textbf{47.34} & \textbf{1299} & \textbf{1566}\\
        \bottomrule
    \end{tabular}
    \label{tab:agent_performance}
\end{table}



We used Welch’s t-test \cite{welch_test} to compare the \textit{SoftGNN}$_{90\%}\, N-1$ agent against \gls{greedy}, \textit{HardGNN}$_{90\%}$, \textit{SoftFNN}$_{90\%}$, and \textit{TopoAgentFix}$_{85-95\%}$, and in all cases rejected the null hypothesis ($p<0.05$; see Tab. \ref{tab:significance}), indicating significant differences. D’Agostino’s test\cite{dagostino_test} confirmed that the data adhered to normality.

\begin{table}[h]
\centering
\caption{Test Results of the Welch’s t-test~\cite{welch_test} with the hypothesis $H_0 : \mu_i = \mu_j$ against the alternative hypothesis $H_1 : \mu_i \ne \mu_j$.}
\label{tab:significance}
\renewcommand{\arraystretch}{1.2}
\setlength{\tabcolsep}{8pt}   

\begin{tabular}{@{}l|l@{}}
\toprule
\textbf{$H_0$ Hypothesis} & \textbf{p-value} \\
\midrule
$H_0$ : $\mu_{\textit{Greedy}_{90\%}}$ =  $\mu_{\textit{SoftGNN}_{90\%}\, N-1}$ & $1.4 \times 10^{-6}$  \\
$H_0$ : $\mu_{\textit{HardGNN}_{90\%}}$ =  $\mu_{\textit{SoftGNN}_{90\%}\, N-1}$ & $1.6 \times 10^{-6}$  \\
$H_0$ : $\mu_{\textit{SoftFNN}_{90\%}}$ =  $\mu_{\textit{SoftGNN}_{90\%}\, N-1}$ & $0.003$  \\
$H_0$ : $\mu_{\textit{TopoAgent}_{85-95\%}}$ =  $\mu_{\textit{SoftGNN}_{90\%}\, N-1}$ & $0.013$ \\
\bottomrule
\end{tabular}
\end{table}

These results underscore three key trends: (1) Soft-label \gls{il} significantly outperforms hard-label \gls{il}, (2) \glspl{gnn} exploit the grid topology to improve decision-making, and (3) post-hoc N-1 evaluation further elevates performance by prioritizing N-1 resilient actions.

\section{Discussion}
The experimental results demonstrate that our soft-label imitation learning (IL) approach, which leverages soft scores over viable topology actions, consistently outperforms both hard-label \gls{il} methods and the expert agent itself. This section synthesizes the key insights and contextualizes them within the landscape of  \gls{il}.

Conventional hard-label \gls{il} methods inherit and amplify the flaws of the expert by enforcing rigid, deterministic policies. Our results show that both the \textit{HardFNN}$_{90\%}$ and \textit{HardGNN}$_{90\%}$ agents are on par with the \gls{greedy} agent while performing significantly worse than their soft-label counterparts. This gap underscores how hard labels propagate the expert’s biases, such as favoring in-optimal actions that might mitigate overloads for singular grid states but lead to unstable topologies for following states and hence disrupt long-term performance.

In contrast, soft labels enable the agent to generalize across states by learning structural patterns in the action space rather than memorizing individual decisions. By learning from soft scores, the model observes which actions are effective for each scenario and infer which actions are universally effective. This also reduces the label noise and the chance of overfitting to singular sub-optimal actions. We argue that the soft labels work like a confidence score, where the confidence decreases whenever there are many viable actions. This avoids overfitting to the action with the highest -- yet not far off -- score. Especially in these low confidence situations where the scores are distributed among multiple effective actions, the exact order of the actions is less important for the model to predict. This is consistent with the use of the KL divergence loss which doesn't account for the order of the predictions but rather the element-wise deviations from the target. Hence, the overall viability of actions is assigned more importance than the exact order according to the labels. This fits our use case perfectly, since it is merely important to bring line loads below a certain threshold. Because the model is able to observe the richer soft label, it is able to assess the general effectiveness of actions, resulting in a tendency to rank actions higher that contribute to reducing line loads in the training set more frequently.

It is even desired to output multiple action recommendations, which can be evaluated with more scrutiny, such as to their impact on the N-1 load flows. Our \textit{SoftGNN}$_{90\%}\, N-1$ agent does exactly that for the top 10 actions and achieves a higher score and longer survival time.  The flexibility of having multiple recommendations for operators is critical for power grids, where topology optimization must not only reduce line load, but also consider other optimization tasks, such as N-1 security or the topology depth \cite{Viebahn24}.

Bridging our results to recent advances in \gls{il}, \cite{Wu_imperfect} demonstrated that confidence scores which indicates the quality of demonstrations enable \gls{il} agents to recover optimal policies from imperfect data. Similarly, our soft labels can be interpreted as such confidence scores. They show that reweighting imperfect demonstrations using confidence signals improves policy robustness. By borrowing this principle to power grid control, we show that soft labels surpass the performance of the imperfect expert by synthesizing a richer understanding of effective actions. By encoding uncertainty through soft labels, the agent avoids over-committing to suboptimal decisions and is therefore more robust to unseen grid states. It's noteworthy to point that the same phenomenon, of students models outperforming expert models as agents, was also confirmed by \cite{dejong2025}.

The integration of GNNs amplifies the benefits of soft labels by explicitly modeling the topological structure of the power grid. While the soft-label \gls{fnn} agent improved performance over its hard-label counterpart, adding a \gls{gnn} achieved the highest scores and median survival times. \glspl{gnn} enhance decision-making by propagating congestion reduction strategies across interconnected substations and lines, ensuring that physical grid constraints are considered. This result is consistent with recent studies \cite{grlsurvey2024,dejong2025}.

Applying \gls{ml} techniques in real-world control rooms as decision support for (topological) remedial actions is still in its infancy. For example, the GridOptions tool \cite{viebahn2024gridoptions} is one of the first AI-based decision-support tools deployed in a \gls{tso} control room. However, the scope of the first version of the GridOptions tool has been limited in several ways \cite{leyliabadi2025}. In particular, the optimization approach does not exploit \gls{ml} yet, and, hence, is slow and inflexible. Consequently, our method’s success has direct relevance to real-world grid operations. By training on diverse action soft scores, the agent becomes resilient to unexpected grid disturbances, e.g., equipment failures or renewable volatility. Our approach maximizes the utility of topology actions by identifying high-impact reconfigurations and therefore reduces the need for costly redispatch. Since we deal with critical infrastructure, our system is developed solely as a decision-support tool that provides action recommendations while leaving the final decision-making authority to qualified human operators. 

While our approach marks a significant advancement, several challenges remain. Future work could explore action sampling strategies as well hybrid approaches applying \gls{rl} fine-tuning to \gls{il} models to capture true long-term dependencies. Scaling to real-world sized power grids with different topologies will validate the method’s broader applicability.

\section{Conclusion}

In this study, we introduce a novel imitation learning framework that leverages soft labels -- derived from comprehensive load flow simulations -- to capture multiple effective topology actions in power grid control. Our approach overcomes the limitations of traditional hard-label methods, which tend to rigidly follow a single expert decision and propagate its biases. By integrating graph neural networks, our agent learns to capture the grid’s inherent spatial structure, leading to an improved performance. The impact of the proposed agent was studied on a benchmark IEEE 118-Bus transmission system. We find that the proposed method outperforms state-of-the-art RL agents and the greedy expert itself, showing the potential of soft-label imitation learning. 

\section{Acknowledgements}
This work was supported by: (i) AI4REALNET founded by the European Union’s Horizon Europe Research and Innovation program under the Grant Agreement No 101119527. Views and opinions expressed are however those of the author(s) only and do not necessarily reflect those of the European Union. Neither the European Union nor the granting authority can be held responsible for them. (ii) Graph Neural Networks for Grid Control (GNN4GC) founded by the Federal Ministry for Economic Affairs and Climate Action Germany(03EI6117A).(iii) Reinforcement Learning for Cognitive Energy Systems (RL4CES) founded by the German Federal Ministry of Education and Research (01|S22063B).

\bibliographystyle{splncs04}
\bibliography{biblio}

\end{document}